\documentclass[11pt,a4paper]{article}
\usepackage[hyperref]{emnlp2020}
\usepackage{times}
\usepackage{latexsym}

\usepackage[utf8]{inputenc}
\usepackage{graphicx}
\usepackage{pgfplots}

\usepackage{microtype}
\usepackage{amsmath}
\usepackage{amssymb}

\usepackage{graphicx}
\usepackage{comment}
\usepackage{array}
\usepackage{mathtools}

\aclfinalcopy % Uncomment this line for the final submission
\def\aclpaperid{2697} %  Enter the acl Paper ID here

\title{Graph Transformer Networks with Syntactic and Semantic Structures \\ for Event Argument Extraction}

\author{ Amir Pouran Ben Veyseh\textsuperscript{\rm 1}, Tuan Ngo Nguyen\textsuperscript{\rm 1} and Thien Huu Nguyen\textsuperscript{\rm 1,2} \\
\textsuperscript{\rm 1} Department of Computer and Information Science, University of Oregon,
\\Eugene, OR 97403, USA\\
\textsuperscript{\rm 2} VinAI Research, Vietnam\\
  \texttt{\{apouranb,tnguyen,thien\}@cs.uoregon.edu} \\
}

\date{}

\begin{document}
\maketitle
\begin{abstract}
The goal of Event Argument Extraction (EAE) is to find the role of each entity mention for a given event trigger word. It has been shown in the previous works that the syntactic structures of the sentences are helpful for the deep learning models for EAE. However, a major problem in such prior works is that they fail to exploit the semantic structures of the sentences to induce effective representations for EAE. Consequently, in this work, we propose a novel model for EAE that exploits both syntactic and semantic structures of the sentences with the Graph Transformer Networks (GTNs) to learn more effective sentence structures for EAE. In addition, we introduce a novel inductive bias based on information bottleneck to improve generalization of the EAE models. Extensive experiments are performed to demonstrate the benefits of the proposed model, leading to state-of-the-art performance for EAE on standard datasets.

%. GTNs offer an mechanism to combine syntactic and semantic structures and infer more effective sentence structures for EAE.

%and achieve the state-of-the-art performance on standard datasets.

%We achieve the state-of-the-art performance for EAE on standard datasets.

%Our extensive experiments demonstrate the benefits of the proposed model and achieve the state-of-the-art performance on standard datasets.

%However, they fail to effectively model the entire graph structure and also they ignore the semantic relations between words. In order to address these issues, in this paper, we propose to employ the new architecture of Graph Transformer Network (GTN) to effectively employ the syntactic and semantic structure of the sentence and model the interaction between them. Moreover, we introduce two new inductive bias in our model to rectify the inherent limitations of the original GTN architecture. Our extensive experiments on two benchmark datasets prove the effectiveness of the proposed model for the task of EAE, leading to new state-of-the-art results on both datasets. 
\end{abstract}

\section{Introduction}

%\footnote{\tiny \url{https://www.ldc.upenn.edu/sites/www.ldc.upenn.edu/files/english-events-guidelines-v5.4.3.pdf}}

Event Extraction (EE) is an important task of Information Extraction that aims to recognize events and their arguments in text. In the literature, EE is often divided into two sub-tasks: (1) Event Detection (ED) to detect the event trigger words, and (2) Event Argument Extraction (EAE) to identity the event arguments and their roles for the given event triggers. In recent years, ED has been studied extensively with deep learning while EAE is relatively less explored \cite{wang2019hmeae}. As EAE is necessary to accomplish EE and helpful for many downstream applications \cite{yang2003structured,cheng2018implicit}, further studies are required to improve the performance of EAE. This work focuses on EAE to meet this requirement for EE.

The current state-of-the-art methods for EAE have involved deep learning models that compute an abstract representation vector for each word in the input sentences based on the information from the other context words. The representation vectors for the words are then aggregated to perform EAE \citep{chen2015event,nguyen2016joint}. Our main motivation in this work is to exploit different structures in the input sentences to improve the representation vectors for the words in the deep learning models for EAE. In this work, a sentence structure (or view) refers to an importance score matrix whose cells quantify the contribution of a context word for the representation vector computation of the current word for EAE. In particular, we consider two types of sentence structures in this work, i.e., syntactic and semantic structures. As such, the importance score for a pair of words in the syntactic structures is determined by the syntactic connections of the words in the dependency parsing trees while the contextual semantics of the words in the input sentences are exploited to compute the importance scores in the semantic structures. Consider the following sentence as an example:

%the syntactic structures (i.e., the dependency parsing trees) determine the importance score for a pair of words based on the syntactic connections of the words in the parsing trees. he semantic structures, the importance score for a pair of words is conditioned on the contextual semantics of the words in the input sentence.

%For instance, in the sentence ``{\it A reporter who had been lost for three months is found dead this morning.}'' with the trigger word ``{\it dead}'' (of type {\it Die}) and the entity mention ``{\it reporter}'', the word ``{\it found}'' is the most important word to reveal the {\it Victim} role of ``{\it reporter}'' for the event triggered by ``{\it dead}''. Using the dependency tree for this sentence, the direct links between ``{\it found}'' and the two words ``{\it reporter}'' and ``{\it dead}'' can be leveraged to encourage the deep learning models to focus more on the context word ``{\it found}'' in the representation computation for ``{\it reporter}'' and ``{\it dead}''. This will lead to the improved representation vectors for ``{\it reporter}'' and ``{\it dead}'' for better argument prediction in EAE. In this work, several variants of the syntactic structures are proposed to enrich the word representations for EAE based on dependency trees. Second, for the semantic structures, the importance score for a pair of words is conditioned on the contextual semantics of the words in the input sentence. Consider the following sentence as an example:

{\it Iraqi Press constantly report interviews with \underline{\textbf{Hussain Molem}}, the \textbf{Hanif Bashir's son-in-law}, while US officials confirmed \textbf{all Bashir's family members} were \underline{\textbf{killed}} last week.}

In this sentence, an EAE system should be able to realize the entity mention ``{\it Hussain Molem}'' as the {\it Victim} of the {\it Attack} event triggered by ``{\it killed}''. As ``{\it Hussain Molem}'' and ``{\it killed}'' are far away from each other in the sentence as well as its dependency tree, the EAE models might find it challenging to make the correct prediction in this case. In order for the models to be successful in this case, our intuition is that the models should first rely on the direct connections between ``{\it killed}'' and ``{\it all Bashir's family members}'' in the dependency tree to capture the role of ``{\it all Bashir's family members}'' in the representation vectors for ``{\it killed}''. Afterward, the models can rely on the close semantic similarity between ``{\it all Bashir's family members}'' and ``{\it the Hanif Bashir's son-in-law}'' to further connect ``{\it the Hanif Bashir's son-in-law}'' to ``{\it killed}'' so the role information of ``{\it the Hanif Bashir's son-in-law}'' can be recorded in the representation vector for ``{\it killed}''. Finally, the direct apposition relation between ``{\it the Hanif Bashir's son-in-law}'' and ``{\it Hussain Molem}'' can be exploited to connect ``{\it Hussain Molem}'' with ``{\it killed}'' to obtain the necessary representations to perform argument prediction for ``{\it Hussain Molem}''. On the one hand, this example suggests that both syntactic and semantic structures are necessary for the EAE models. On the other hand, the example also hints that it is not enough to apply the syntactic and semantic structures separately. In fact, these structures should be explicitly combined to complement each other on identifying important context words to obtain effective representations for EAE. To our knowledge, this is the first work to explore syntactic and semantic structures for EAE.

How should we combine the syntactic and semantic structures to aid the learning of effective representations for EAE? In this work, we propose to employ Graph Transformer Networks (GTN) \cite{yun2019graph} to perform the syntax-semantic merging for EAE. GTNs facilitate the combination of multiple input structures via two steps. The first step obtains the weighted sums of the input structures, serving as the intermediate structures that are able to capture the information from different input perspectives (i.e., structures). In the second step, the intermediate structures are multiplied to generate the final structures whose goal is to leverage the multi-hop paths/connections between a pair of nodes/words (i.e., involving the other words) to compute the importance score for the final structures. As the multi-hop paths with heterogeneous types of connections along the way (i.e., syntax or semantic) has been illustrated to be helpful in our running example (i.e., between `{\it Hussain Molem}'' and `{\it killed}''), we expect that GTNs can help to combine the syntactic and semantic structures to produce effective representations for EAE.

Finally, in order to further boost the performance for EAE, we propose a novel inductive bias for the proposed GTN model, aiming to improve the generalization of GTNs using the Information Bottleneck idea \citep{tishby2000information,Belghazi:18}. In particular, the use of the rich combined structures from syntax and semantics might augment GTNs with high capacity to encode the detailed information in the input sentences. Coupled with the generally small training datasets for EAE, the GTN models could learn to preserve all the context information in the input sentences, including the irrelevant information for EAE. This likely leads to the overfitting of GTNs on the training data. In order to overcome this issue, we propose to treat the GTN model in this work as an information bottleneck in which the produced representations of GTNs are trained to not only achieve good prediction performance for EAE but also minimize the mutual information with the input sentences \citep{Belghazi:18}. To this end, we introduce the mutual information between the generated representations of GTNs and the input sentences as an additional term in the overall loss function to improve the generalization of GTNs for EAE. Our extensive experiments on two benchmark datasets for EAE show that the proposed model can achieve the state-of-the-art performance for EAE.

\section{Related Work}

EAE is one of the two subtasks in EE (the other one is ED) that has been approached early by the feature-based models \citep{ahn2006stages,ji2008refining,patwardhan2009unified,liao2010filtered,liao2010using,Riedel:11,Hong:11,McClosky:11,li2013joint,Makoto:14,Yang:16}. The recent work on EE has focused on deep learning to improve the models' performance \citep{chen2015event,sha2018jointly,Zhang:19,Yang:19,Nguyen:19,Zhang:20}. Among the two subtasks of EE, while ED has been studied extensively by the recent deep learning work \citep{Nguyen:15,chen2015event,Nguyen:16g,Chen:17,Liu:17,Liu:18,zhao2018document,Wang:19,Lai:20c}, EAE has been relatively less explored. The closest work to ours is \citep{wang2019hmeae} that focuses on EAE and exploits the concept hierarchy of event argument roles to perform the task. Our work differs from \citep{wang2019hmeae} in that we employ the syntactic and semantic structures of the sentences to better learn the representations for EAE. We also note some new directions on EE based on zero-shot learning \citep{Huang:18}, few-shot learning \cite{Lai:20a,Lai:20b} and multimodal learning \citep{Zhang:17}.

\section{Model}

EAE can be formulated as a multi-class classification problem in which the input involves a sentence $W=w_1,w_2,\ldots,w_N$ ($w_i$ is the $i$-th word/token in the sentence of length $N$), and an argument candidate and event trigger at indexes $a$ and $t$ in the sentence (i.e., the words $w_a$ and $w_e$) respectively. The goal in this problem is to predict the role that the argument candidate $w_a$ plays in the event triggered by $w_e$. Note that the set of the possible roles also include a special type {\it None} to indicate that the argument candidate is not an actual argument of the event (i.e., no roles). In order to achieve a fair comparison, following the prior work on EAE \cite{wang2019hmeae}, we take as inputs the event triggers that are detected by an independent model (i.e., the model in \cite{Wang:19}) and separate from our proposed model. We also consider each entity mention in the sentence as a candidate argument for the role prediction task. Our model for EAE in this work involves four major components: (i) sentence encoding, (ii) structure generation, (iii) structure combination, and (iv) model regularization as described in the following.

%We will describe these components in the following sections.

%In order to solve EAE, our model in this work involves four major components: (i) sentence encoding to encode the input sentence, (ii) structure generation to compute the syntactic and semantic structures for the input, (iii) structure combination to mix the generated structures with GTNs, and (iv) model regularization to improve the performance of GTNs. We will describe these components in the following sections.

%The EAE problem can be formulated as a multi-class classification problem. Formally, given an input sentence $W=w_1,w_2,\ldots,w_N$ where $w_i$ is the $i$-th word in the sentence $W$ of length $N$, and an entity mention and event trigger at indexes $a$ and $t$, our goal is to predict the role of the argument $w_a$ in the event $w_e$ in $W$. Following previous work, the event trigger $w_e$ is detected by an independent model separate from our proposed model \cite{wang2019hmeae}. Also, each entity mention in $W$ is a candidate argument.

\subsection{Sentence Encoding}
\label{sec:wordrep}
%To represent each word $w_i$, following previous work, we encode each word into a dense vector $x_i$. This dense vector is the concatenation of the following vectors: 

To represent the sentence, we encode each word $w_i$ with a real-valued vector $x_i$ that is the concatenation of the two following vectors: (i) the embeddings of the relative distances of the word $w_i$ to the argument candidate (i.e., $i-a$) and event trigger (i.e., $i-e$) (these embeddings are initialized randomly and updated during training), and (ii) the BERT embedding of the word $w_i$. In particular, to achieve a fair comparison with \citep{wang2019hmeae}, we run the BERT base cased model \cite{Devlin:19} over $W$ and use the hidden vector for the first wordpiece of $w_i$ in the last layer of BERT as the embedding vector (of 768 dimensions) for $w_i$. 

%Note that BERT has been shown to achieve the state-of-the-art performance for EAE in the prior work \cite{wang2019hmeae}.

%\begin{itemize}
    %\item POS tag embedding of the word $w_i$
    %\item entity type embedding of the word $w_i$
%    \item 
%(i) The embeddings of the relative distances of the word $w_i$ to the argument candidate (i.e., $i-a$) and event trigger (i.e., $i-e$). These embeddings are initialized randomly and updated during training.
    %\item 
    
%(ii) The BERT embedding of the word $w_i$. In particular, we run the BERT model in \cite{Devlin:19} over the input sentence $W$ and use the hidden vector for the first wordpiece of $w_i$ in the last BERT layer as the embedding for $w_i$. Note that BERT embedding has been shown to achieve the state-of-the-art performance for EAE in the prior work \cite{wang2019hmeae}.
%\end{itemize}

The word encoding step then produces a sequence of vectors $X=x_1,\ldots,x_N$ to represent the input sentence $W$. In order to better combine the BERT embeddings and the relative distance embeddings, we further feed $X$ into a Bidirectional Long-short Term Memory network (BiLSTM), resulting in the hidden vector sequence $H = h_1,\ldots,h_N$ as the representation vectors for the next steps.

%After encoding all words into vectors $X=x_1,x_2,...,x_N$, the representations vectors are fed into the subsequent neural components. Our model is consists of the following components: (1) View Extractor: This components extracts semantic and various syntactic structure out of the given sentence, (2) View Combiner: This components employ GTN to combines views extracted in the previous component, (3) Regularizer: This component regularizes the representations obtained from the combined views from the previous section to increase the diversity of the modeled structures and prevent overfitting using Information Bottleneck and (4) Prediction: This component exploits the regularized representations from the previous components to make prediction about the role of the given argument in the event mentioned in the sentence $W$. The rest of this section will provide the details of each of these components. 

%\subsection{Extracting Views}

\subsection{Structure Generation}
\label{sec:gen}
As presented in the introduction, the motivation for our EAE model is to employ the sentence structures to guide the computation of effective representation vectors for EAE with deep learning. These sentence structures would involve the score matrices of size $N \times N$ in which the score at the $(i,j)$ cell is expected to capture the importance of the contextual information from $w_j$ with respect to the representation vector computation of $w_i$ in the deep learning models for EAE (called the importance score for the pair $(w_i,w_j)$). In this work, we consider two types of sentence structures for EAE, i.e., the syntactic structures and the semantic structures.

{\bf Syntactic Structures}: It has been shown in the prior work that the dependency relations in the dependency trees can help to connect a word to its important context words to obtain effective representation vectors for EAE \cite{sha2018jointly}. To this end, we use the adjacency matrix $A^d$ of the dependency tree for $W$ as one of the syntactic structures for EAE in this work. Note that $A^d$ here is a binary matrix whose cell $(i,j)$ is only set to 1 if $w_i$ and $w_j$ are linked in the dependency tree for $W$.

One problem with the $A^d$ structure is that it is agnostic to the argument candidate $w_a$ and event trigger $w_e$ for our EAE task. As the argument candidate and event trigger are the most important words in EAE, we argue that the sentence structures should be customized for those words to produce more effective representation vectors in $W$ for EAE. In order to obtain the task-specific syntactic structures for EAE, our intuition is that the closer words to the argument candidate $w_a$ and event trigger $w_e$ in the dependency tree would be more informative to reveal the contextual semantics of $w_a$ and $w_e$ than the farther ones in $W$ \citep{Nguyen:18}. These syntactic neighboring words of $w_a$ and $w_e$ should thus be assigned with higher importance scores in the sentence structures, serving as the mechanism to tailor the syntactic structures for the argument candidate and event trigger for EAE in this work. Consequently, besides the general structure $A^d$, we propose to generate two additional customized syntactic structures for EAE based on the lengths of the paths between $w_a$, $w_e$ and the other words in the dependency tree of $W$ for EAE (i.e., one for the argument candidate and one for the event trigger). In particular, for the argument candidate $w_a$, we first compute the length $d^a_i$ of the path from $w_a$ to every other word $w_i$ in $W$. The length $d^a_i$ is then converted to an embedding vector $\hat{d}^a_i$ by looking up a length embedding table $D$ (initialized randomly and updated during training): $\hat{d}^a_i = D[d^a_i]$. Afterward, we generate the argument-specific syntactic structure $A^a = \{s^a_{i,j}\}_{i,j=1..N}$ via: $s^a_{i,j} = \textit{sigmoid}(FF([\hat{d}^a_i, \hat{d}^a_j, \hat{d}^a_i \odot \hat{d}^a_j, |\hat{d}^a_i - \hat{d}^a_j|, \hat{d}^a_i + \hat{d}^a_j]))$ where $[]$ is the vector concatenation, $\odot$ is the element-wise multiplication, and $FF$ is a two-layer feed-forward network to convert a vector to a scalar. We expect that learning the syntactic structures in this way would introduce the flexibility to infer effective importance scores for EAE. 

The same procedure can then be applied to generate the trigger-specific syntactic structure $A^e = \{s^e_{i,j}\}_{i,j=1..N}$ for the event trigger $w_e$ for the EAE model in this work. Finally, the general structure $A^d$ and the task-specific syntactic structures $A^a$ and $A^e$ would be used as the syntactic structures for our structure combination component in the next step.

%, and $\sigma$ is the {\it sigmoid} function.

%\begin{equation}
%\small
%    s^a_{i,j} = \sigma(FF([\hat{d}^a_i, \hat{d}^a_j, \hat{d}^a_i * \hat{d}^a_j, |\hat{d}^a_i - \hat{d}^a_j|, \hat{d}^a_i + \hat{d}^a_j]))
%\end{equation}

{\bf Semantic Structure} The semantic structure aims to learn the importance score for a pair of words $(w_i,w_j)$ by exploiting the contextual semantics of $w_i$ and $w_j$ in the sentence. As mentioned in the introduction, the semantic structure is expected to provide complementary information to the syntactic structures, that once combined, can lead to effective representation vectors for EAE. In particular, we employ the Bi-LSTM hidden vectors $H=h_1,\ldots,h_N$ in the sentence encoding section to capture the contextual semantics of the words for the semantic structure in this work. The semantic importance scores $s^s_{i,j}$ for the semantic structure $A^s = \{s^s_{i,j}\}_{i,j=1..N}$ can then be learned via $s^s_{i,j} = f(h_i,h_j)$ where $f$ is some function to produce a score for $h_i$ and $h_j$. Motivated by the self-attention scores in \citep{vaswani2017attention}, we use the following key and query-based function for the $f$ function for $s^s_{i,j}$: 
\begin{equation}
%\small
    \begin{split}
        k_i & = U_k h_i, q_i = U_q h_i \\
        s^s_{i,j} & = \exp(k_i q_j) / \sum_{v=1..N} \exp(k_i q_v)
    \end{split}
    \label{eq:sem}
\end{equation}
where $U_k$ and $U_q$ are trainable weight matrices, and the biases are omitted in this work for brevity.

%$k_i = U_k h_i$, $q_i = U_q h_i$, $s^s_{i,j} = \exp(k_i q_j) / \sum_{v=1..N} \exp(k_i q_v)$ where $U_k$, $U_q$ and $U_v$ are trainable weight matrices and the biases are omitted in this work for brevity.

Similar to the general syntactic structure $A^d$, a problem for this function is that the semantic scores $s^s_{i,j}$ are not aware of the argument candidate and the event trigger words, the two important words for EAE. To this end, we propose to involve the contextual semantics of the argument candidate $w_a$ and event trigger $w_e$ (i.e., $h_a$ and $h_t$) in the computation of the semantic structure score $s^s_{i,j}$ for EAE using:
\begin{equation}
%\small
    \begin{split}
        c^k & = \sigma(V_k [h_a,h_t]), k'_i = c^k \odot k_i \\
        c^q & = \sigma(V_q [h_a,h_t]), q'_i = c^q \odot q_i \\
        s^s_{i,j} & = \exp(k'_i q'_j) / \sum_{v=1..N} \exp(k'_i q'_v)
    \end{split}
    \label{eq:semcus}
\end{equation}
The rationale in this formula is to use the hidden vectors for the argument candidate $w_a$ and event trigger $w_e$ to generate the task-specific control vectors $c^k$ and $c^q$. These control vectors are then employed to filter the information in the key and query vectors (i.e., $k_i$ and $q_i$) so only the relevant information about $w_a$ and $w_e$ is preserved in $k_i$ and $q_i$ via the element-wise products $\odot$. The resulting key and query vectors (i.e., $k'_i$ and $q'_i$) would then be utilized to compute the task-specific importance score $s^s_{i,j}$ for the semantic structure $A^s$ in this work.

\subsection{Structure Combination}
\label{sec:combining}
The four initial structures in $\mathcal{A} = [A^d,A^a,A^e,A^s]$ can be interpreted as four different types of relations between the pairs of words in $W$ (i.e., using the syntactic and semantic information) (called the word relation types). The cell $(i,j)$ in each initial structure is deemed to capture the degree of connection between $w_i$ and $w_j$ based on their direct interaction/edge (i.e., the one-hop path $(w_i,w_j)$) and the corresponding relation type for this structure. Given this interpretation, this component seeks to combine the four initial structures in $\mathcal{A}$ to obtain richer sentence structures for EAE. On the one hand, we expect the importance scores between a pair of words $(w_i,w_j)$ in the combined structures to be able to condition on the possible interactions between $w_i$, $w_j$ and the other words in the sentence (i.e., the multi-hop paths between $w_i$ and $w_j$ that involve the other words). On the other hand, the multi-hop paths between $w_i$ and $w_j$ for the importance scores should also be able to accommodate the direct edges/connections between the words of different relation types (i.e., the heterogeneous edge types). Note that both the multi-hop paths and the heterogeneous edge types along the paths (i.e., for syntax and semantics) have been demonstrated to be helpful for EAE in the introduction. Consequently, in this work, we propose to apply Graph Transformer Networks (GTNs) \citep{yun2019graph} to simultaneously achieve these two goals for EAE.

In particular, following \citep{yun2019graph}, we first add the identity matrix $I$ (of size $N \times N$) into the set of structures in $\mathcal{A}$ to enable GTNs to learn the multi-hop paths at different lengths, i.e., $\mathcal{A}=[A^d,A^a,A^e,A^s,I]=[\mathcal{A}_1,\mathcal{A}_2,\mathcal{A}_3,\mathcal{A}_4,\mathcal{A}_5]$. Given the initial structures in $\mathcal{A}$ and inspired by the transformers in \citep{vaswani2017attention}, the GTN model is organized into $C$ channels for which the $i$-the channel involves $M$ intermediate structures $Q^i_1, Q^i_2, \ldots, Q^i_M$ of size $N \times N$ ($1 \le i \le C$) (i.e., corresponding to $M-1$ layers in GTNs). In GTNs, each intermediate structure $Q^i_j$ is computed via a weighted sum of the individual input structures with learnable weights $\alpha^i_{j,v}$: $Q^i_j = \sum_{v=1..5}\alpha^i_{j,v} \mathcal{A}_v$. The weighted sums enable the intermediate structures to reason with any of the four initial word relation types depending on the context, thus offering the structure flexibility for the model. Afterward, in order to capture the multi-hop paths for the importance scores in the $i$-th channel, the intermediate structures are multiplied to obtain a single sentence structure $Q^i$ for this channel: $Q^i = Q^i_1 \times Q^i_2 \times \ldots \times Q^i_M$ (called the final structures). It has been shown in \citep{yun2019graph} that $Q^i$ is able to model any multi-hop paths between the words with lengths up to $M$. Such multi-hop paths can also host heterogeneous word relation types in the edges (due to the flexibility of the intermediate structures $Q^i_j$), thus introducing rich sentence structures for our EAE problem.

%The weighted sums in the intermediate structures serve as the main mechanism to combine the syntactic and semantic structures for our EAE problem in this work. Afterward, in order to capture the multi-hop paths for the importance scores in the $i$-th layer, the intermediate structures are multiplied to obtain a single sentence structure $Q^i$ for the $i$-th GTN layer: $Q^i = Q^i_1 \times Q^i_2 \times \ldots \times Q^i_M$ (called the final structures). It has been shown in \citep{yun2019graph} that $Q^i$ is able to model any multi-hop paths between the words with lengths up to $M$, introducing a rich sentence structure for our EAE problem.

%Note that the set of the final structures in different layers of the GTN model (i.e., $Q^1,Q^2,\ldots,Q^L$) is inspired by the multiple self-attention heads in the transformer model in \citep{vaswani2017attention} that aim to facilitate the emergence of different semantic aspects (i.e., one for each layer) in the model.

In the next step, the final structures $Q^1,Q^2,\ldots,Q^C$ of GTN are treated as different adjacency matrices for the fully connected graph between the words in $W$. These matrices would then be consumed by a Graph Convolutional Network (GCN) model \citep{kipf2016semi} to produce more abstract representation vectors for the words in our EAE problem. In particular, the GCN model in this work consists of several layers (i.e., $G$ layers in our case) to compute the representation vectors at different abstract levels for the words in $W$. For the $k$-th final structure $Q^k$, the representation vector for the word $w_i$ in the $t$-th GCN layer would be computed via:
\begin{equation}
%\small
    \bar{h}^{k,t}_i = ReLU(U^t \sum_{j=1..N}  \frac{Q^k_{i,j}  \bar{h}^{k,t-1}_j}{\sum_{u=1..N}Q^k_{i,u}}) 
    \label{eq:gcn}
\end{equation}
where $U^t$ is the weight matrix for the $t$-th GCN layer and the input vectors $\hat{h}^{k,0}_i$ for the GCN model are obtained from the Bi-LSTM hidden vectors (i.e., $\hat{h}^{k,0}_i = h_i$ for all $1 \le k \le C$, $1\le i \le N$).

Given the outputs from the GCN model, the hidden vectors in the last GCN layer (i.e., the $G$-th layer) of the word $w_i$ for all the final structures (i.e., $\bar{h}^{1,G}_i, \bar{h}^{2,G}_i, \ldots, \bar{h}^{C,G}_i$) are then concatenated to form the final representation vector $h'_i$ for $w_i$ in the proposed GTN model: $h'_i = [\bar{h}^{1,G}_i,\bar{h}^{2,G}_i,\ldots,\bar{h}^{C,G}_i]$. Finally, in order to predict the argument role for $w_a$ and $w_e$ in $W$, we assemble a representation vector $R$ based on the hidden vectors for $w_a$ and $w_e$ from the GCN model via: $R = [h'_a, h'_e, MaxPool(h'_1,h'_2,\ldots,h'_N)]$. This vector is then sent to a two-layer feed-forward network with softmax in the end to produce a probability distribution $P(.|W,a,t)$ over the possible argument roles. We would then optimize the negative log-likelihood $L_{pred}$ to train the model in this work: $\mathcal{L}_{pred} = -P(y|W,a,t)$ where $y$ is the golden argument role for the input example.

\subsection{Model Regularization}
\label{sec:reg}

As presented in the introduction, the high representation learning capacity of the GTN model could lead to memorizing the information that is only specific to the training data (i.e., overfitting). In order to improve the generalization, we propose to regularize the representation vectors obtained by GTN so only the effective information for EAE is preserved in the GTN representations for argument prediction and the nuisance information of the training data (i.e., the irrelevant one for EAE) can be avoided. To this end, we propose to treat the GTN model as an Information Bottleneck (IB) \cite{tishby2000information} in which the GTN-produced representation vectors $H' = h'_1, h'_2, \ldots, h'_N$ would be trained to both (1) retain the effective information to perform argument prediction for EAE (i.e., the high prediction capacity) and (2) maintain a small Mutual Information (MI)\footnote{In information theory, MI evaluates how much information we know about one random variable if the value of another variable is revealed.} with the representation vectors of the words from the earlier layers of the model (i.e., the minimality of the representations) \citep{Belghazi:18}. In this work, we follow the common practice to accomplish the high prediction capacity by using the GTN representation vectors to predict the argument roles and minimizing the induced negative log-likelihood in the training phase. However, for the minimality of the representations, we propose to achieve this by explicitly minimizing the MI between the GTN-produced vectors $H' = h'_1, h'_2, \ldots, h'_N$ and the BiLSTM hidden vectors $H = h_1, h_2, \ldots, h_N$ from sentence encoding. By encouraging a small MI between $H$ and $H'$, we expect that only the relevant information for EAE in $H$ is passed through the GTN bottleneck to be retained in $H'$ for better generalization.

%we expect that the small IB between $H$ and $H'$ can help to filter out the extraneous information from the sentence in $H$ and only preserve the relevant information for EAE in $H'$ to improve the overall generalization of the proposed model.

As $H$ and $H'$ are sequences of vectors, we first transform them into the summarized vectors $h$ and $h'$ (respectively) to facilitate the MI estimation via the max-pooling function: $h = MaxPool(h_1,h_2,\ldots,h_N)$ and $h' = MaxPool(h'_1, h'_2, \ldots, h'_N)$. Afterward, we seek to compute the MI between $h$ and $h'$ and include it in the overall loss function for minimization. However, as $h$ and $h'$ are high-dimensional vectors, their MI estimation is prohibitively expensive in this case. To this end, we propose to apply the mutual information neural estimation (MINE) method in \citep{Belghazi:18} to approximate the MI with its lower bound. In particular, motivated by \citep{Hjelm:19}, we propose to further approximate the lower bound of the MI between $h$ and $h'$ via the adversarial approach using the loss function of a variable discriminator. As the MI between $h$ and $h'$ is defined as the KL divergence between the joint and marginal distributions of these two variables, the discriminator aims to differentiate the vectors that are sampled from the joint distribution and those from the product of the marginal distributions for $h$ and $h'$. In our case, we sample from the joint distribution for $h$ and $h'$ by simply concatenating the two vectors (i.e., $[h',h]$) and treat it as the positive example. To obtain the sample from the product of the marginal distribution, we concatenate $h'$ with $\hat{h}$ that is the aggregated vector (via max-pooling) of the BiLSTM hidden vectors for another sentence (obtained from the same batch with the current sentence during training) (i.e., $[h',\hat{h}]$ as the negative example). These positive and negative examples are then fed into a two-layer feed-forward network $D$ (i.e., the discriminator) to produce a scalar score, serving as the probability to perform a binary classification for the variables. Afterward, the logistic loss $\mathcal{L}_{disc}$ of $D$ is proposed as an estimation of the MI between $h$ and $h'$ and added into the overall loss function for minimization: $\mathcal{L}_{disc} = \log (1 + e^{(1-D([h',h]))}) + \log (1 + e^{D([h',\hat{h}])})$.

Finally, the overall loss function $\mathcal{L}$ to train the model in this work would be: $\mathcal{L} = \mathcal{L}_{pred} + \alpha_{disc} \mathcal{L}_{disc}$ where $\alpha_{disc}$ is a trade-off parameter.

\section{Experiments}
\label{sec:exp}
%\subsection{Dataset \& Parameters}

{\bf Datasets \& Parameters}: We evaluate the models on two benchmark datasets, i.e., ACE 2005 and TAC KBP 2016 \citep{ellis2016overview}. ACE 2005 is a widely used EE dataset, involving 599 documents, 33 event subtypes and 35 argument roles. We use the same data split with the prior work \cite{chen2015event,wang2019hmeae} for a fair comparison (i.e., 40 documents for the test data, 30 other documents for the development set, and the remaining 529 documents for the training data). For TAC KBP 2016, as no training data is provided, following \cite{wang2019hmeae}, we use the ACE 2005 training data to train the models and then evaluate them on the TAC KBP 2016 test data. To evaluate the models' performance, for a fair comparison with the previous work \cite{chen2015event,wang2019hmeae}, we consider an argument classification as correct if its predicted event subtype, offsets and argument role match the golden data.  

%To fine-tune the hyper-parameters of the GTN model in this work, we employ the development set of the ACE 2005 dataset. Based on this process, we found the following hyper-parameters: 30 dimensions for POS tag, entity type and distance embedding; BERT$_{base}$ with 768 hidden dimensions for pre-trained word embedding (note that we fix the BERT$_{base}$ parameters during training); 200 hidden dimensions for all feed forward layers; 2 layers of BiLSTM and GCN; 3 layers of GTN; 0.01 for $\beta$; 0.625, 0.125, and 0.250 for $\alpha_{GTN}$, $\alpha_{SA}$ and $\alpha_{BiLSTM}$ (respectively); and 0.1 for trade-off parameters $\alpha_{div}$ and $\alpha_{IB}$ (respectively).

%BERT$_{base}$ with 768 hidden dimensions for the sentence encoding (note that we fix the BERT$_{base}$ parameters during training),

We fine-tune the hyper-parameters for our model on the ACE 2005 development set, leading to the following values: 30 dimensions for the relative distance and length embeddings (i.e., $D$), 200 hidden units for the feed-forward network, BiLSTM and GCN layers, 2 layers for the BiLSTM and GCN modules ($G=2$), $C=3$ channels for GTN with $M=3$ intermediate structures in each channel, and 0.1 for the parameter $\alpha_{disc}$. Finally, besides BERT, we also evaluate the proposed model when BERT is replaced by the {\tt word2vec} embeddings (of 300 dimensions) \citep{Mikolov:13} to make it comparable with some prior works. Note that as in \citep{wang2019hmeae}, the proposed model with BERT takes as inputs the predicted event triggers from the BERT-based ED model in \citep{Wang:19} while the proposed model with {\tt word2vec} utilizes the predicted event triggers from the {\tt word2vec}-based ED model in \citep{chen2015event} for compatibility.

%A reproducibility checklist is shown in Appendix \ref{app:repo}.

% and optimized during training

% for the pre-trained word embeddings

%and 0.1 for both trade-off parameters $\alpha_{div}$ and $\alpha_{disc}$.

%\subsection{Comparison with the state of the art}

{\bf Comparison with the State of the Art}: To evaluate the effectiveness of the proposed model (called {\bf SemSynGTN}), we first compare it with the baselines on the ACE 2005 dataset. Following \citep{wang2019hmeae}, we use the following baselines in our experiments: (i) the feature-based models (i.e., {\bf Li's Joint} \citep{li2013joint} and {\bf RBPB} \citep{sha2016rbpb}), (ii) the deep sequence-based models that run over the sequential order of the words in the sentences (i.e., {\bf DMCNN} \citep{chen2015event}, {\bf JRNN} \citep{nguyen2016joint}, {\bf PLMEE} \citep{Yang:19}, and {\bf DMBERT} (i.e., DMCNN with BERT) \citep{wang2019hmeae}), (iii) the deep structure-based models that employ dependency trees for BiLSTM or GCNs (i.e., {\bf dbRNN} \cite{sha2018jointly} and {\bf JMEE} \citep{Liu:18event}), (iv) the models with Generative Adversarial Imitation Learning ({\bf GAIL} \citep{Zhang:19a}),  and (v) the deep learning model that exploits the hierarchical concept correlation among argument roles (i.e., {\bf HMEAE} \cite{wang2019hmeae}). {\bf HMEAE} is a BERT-based model with the current best EAE performance on ACE 2005.

\begin{table}[ht]
%\small
    \centering
    \resizebox{.46\textwidth}{!}{
    \begin{tabular}{l|c|c|c}
        Model & P & R & F1 \\ \hline
        Li's joint \citep{li2013joint} & 64.7 & 44.4 & 52.7 \\
        RBPB \citep{sha2016rbpb} & 54.1 & 53.5 & 53.8 \\ \hline
        DMCNN \citep{chen2015event} & 62.2 & 46.9 & 53.5 \\ 
        JRNN \citep{nguyen2016joint} & 54.2 & 56.7 & 55.4 \\
        dbRNN \citep{sha2018jointly} & 66.2 & 52.8 & 58.7 \\ 
        GAIL \citep{Zhang:19a} & - & - & 59.7 \\ 
        JMEE \citep{Liu:18event} & 66.8 & 54.9 & 60.3 \\
        \hline
        %SemSynGTN (ours) & 68.4 & 53.9 & {\bf 61.2} \\ \hline \hline
        SemSynGTN (ours) & 68.4 & 55.4 & {\bf 61.2} \\ \hline \hline
        PLMEE \citep{Yang:19}* & 62.3 & 54.2 & 58.0 \\
        DMBERT \citep{wang2019hmeae}* & 58.8 & 55.8 & 57.2 \\
        HMEAE \citep{wang2019hmeae}* & 62.2 & 56.6 & 59.3 \\ \hline
        %SemSynGTN (BERT) (ours)* & 69.3 & 55.4 & \textbf{61.9} 
        SemSynGTN (BERT) (ours)* & 69.3 & 55.9 & \textbf{61.9} 
    \end{tabular}
    }
    \caption{EAE Performance on the ACE 2005 test set. * indicates the models that use BERT.}
    \label{tab:main}
\end{table}

Table \ref{tab:main} presents the performance of the models on the ACE 2005 test set. Note that we distinguish between the models that employ BERT for the pre-trained word embeddings and those that do not for a clear comparison in the table. The most important observation is that the proposed model SemSynGTN significantly outperforms all the baseline models (with $p < 0.01$) no matter if BERT is used as the pre-traind word embeddings or not. SemSynGTN achieves the state-of-the-art performance on ACE 2005 when BERT is applied in the model, thus demonstrating the benefits of the proposed model with the syntactic and semantic structure combination for EAE in this work.

In order to further demonstrate the effectiveness of the proposed model, following the previous work \citep{wang2019hmeae}, we evaluate the models on the TAC KBP 2016 dataset. In particular, we compare SemSynGTN with the top four systems in the TAC KBP 2016 evaluation \citep{dubbin2016improving,hsi016cmu,ferguson2016university}, the DMCNN model in \citep{chen2015event}, and the DMBERT and HMEAE models in \cite{wang2019hmeae}. Note that HMEAE is also the state-of-the-art model for EAE on this dataset. The results are shown in Table \ref{tab:tac} that corroborates our findings from Table \ref{tab:main}. Specifically, SemSynGTN significantly outperforms all the baseline models with large margins ($p < 0.01$) (whether BERT is used or not), thus confirming the advantages of the SemSynGTN model in this work. 

%As SemSynGTN achieves its best performance with BERT, we will always use BERT in the following analysis experiments.

%s which shows the high generalization ability of our model. Moreover, similar to previous experiment, our model achieves the highest precision while it is also close to best recall obtained by HMEAE.

%compare our model with the top systems \cite{dubbin2016improving,hsi016cmu,ferguson2016university} in TAC KBP 2016 competition as well as DMCN \cite{chen2015event} and HMEAE \cite{wang2019hmeae}. The results are demonstrated in Table \ref{tab:tac}. This table corroborates our finding from Table \ref{tab:main}. Specifically, the proposed model significantly outperforms all baselines which shows the high generalization ability of our model. Moreover, similar to previous experiment, our model achieves the highest precision while it is also close to best recall obtained by HMEAE.  

\begin{table}[ht]
%\small
    \centering
    \resizebox{.48\textwidth}{!}{
    \begin{tabular}{l|c|c|c}
        Model & P & R & F1 \\ \hline
        DISCERN-R \citep{dubbin2016improving} & 7.9 & 7.4 & 7.7 \\
        Washington4 \citep{ferguson2016university} & 32.1 & 5.0 & 8.7 \\ 
        CMU CS Event 1 \citep{hsi016cmu} & 31.2 & 4.9 & 8.4 \\
        Washington1 \citep{ferguson2016university} & 26.5 & 6.8 & 10.8 \\
        DMCNN \citep{chen2015event} & 17.9 & 16.0 & 16.9 \\ \hline
        %SemSynGTN (ours) & 39.4 & 15.3 & {\bf 21.2} \\ \hline \hline
        SemSynGTN (ours) & 39.4 & 15.3 & {\bf 22.0} \\ \hline \hline
        DMBERT \citep{wang2019hmeae}* & 22.6 & 24.7 & 23.6 \\
        HMEAE \citep{wang2019hmeae}* & 24.8 & 25.4 & 25.1 \\ \hline
        SemSynGTN (BERT) (ours)* & 51.1 & 19.8 & \textbf{28.5} 
        %SemSynGTN (BERT) (ours)* & 51.1 & 19.8 & \textbf{28.0} 
    \end{tabular}
    }
    \caption{Performance on the TAC KBP 2016 dataset. * indicates the models that use BERT.}
    \label{tab:tac}
\end{table}

%\subsection{Ablation Study}

{\bf Ablation Study}: This part analyzes the effectiveness of the components in the proposed model for EAE by removing each of them from the overall model and evaluating the performance of the remaining models on the ACE 2005 development dataset. In particular, the first major component in SemSynGTN involves structure customization that seeks to tailor the initial syntactic and semantic structures for the argument candidate and event trigger. We evaluate two ablated models for this component: (i) eliminating the task-specific syntactic customization from SemSynGTN that amounts to excluding the customized syntactic structures $A^a$ and $A^e$ from the initial structure set $\mathcal{A}$ (called {\bf SemSynGTN - SynCustom}), and (ii) removing the task-specific semantic customization from SemSynGTN that leads to the use of the simple key-query version (i.e., Equation \ref{eq:sem}) to compute the importance scores in the semantic structure $A^s$ (i.e., instead of using Equation \ref{eq:semcus} as in the full model) (called {\bf SemSynGTN - SemCustom}).

%\footnote{Note that we also re-optimize the number of layers for the GCN model (i.e., $G$), the numbers of intermediate structures and layers for the GTN model (i.e., $M$ and $L$) on the ACE 2005 development set in this case for SemSynGTN - Multi-hop, leading to the values of $G = 2$, $M=3$, and $L = 3$.}. 

The second major component is structure combination that aims to generate the mixed structures from the initial structures in $\mathcal{A}$ via GTN. We consider two ablated models for this component: (i) completely excluding the GTN model and directly applying the GCN model on the initial structures in $\mathcal{A}$ (called {\bf SemSynGTN - GTN}) (the representation vectors from the last GCN layer for the same word with different initial structures are also concatenated in this case), and (ii) keeping the intermediate structures in the channels of the GTN model, but avoiding the intermediate structure multiplications for multi-hop paths in each channel of GTN (called {\bf SemSynGTN - Multi-hop}) (so the GCN model is directly applied over the intermediate structures whose outputs are also concatenated).

%\footnote{Note that for the ablated models in this component, we also re-optimize the number of layers for GCN (i.e., $G$), and the numbers of intermediate structures and channels for GTN (i.e., $M$ and $C$) on the ACE 2005 development set, leading to the values of $G = 2$, $M=3$, and $C = 2$.}

Finally, the third component corresponds to the regularization loss based on information bottleneck $L_{disc}$ in Section \ref{sec:reg}. The removal of $\mathcal{L}_{disc}$ from the overall loss $\mathcal{L}$ leads to the ablated model {\bf SemSynGTN - IB}. As this component relies on the MI between the hidden vectors returned by the BiLSTM and GTN models, we evaluate another variant for SemSynGTN in this case where the regularization loss is eliminated, but the hidden vectors from the BiLSTM model $h_1,h_2,\ldots,h_N$ are included in the final representation vector $R$ for argument role prediction (i.e., $R = [h_a,h_e,h,h'_a,h'_e,h']$) (called {\bf SemSynGTN - IB + LSTM in $R$}). Table \ref{tab:ablation} provides the performance of the ablated models on the ACE 2005 development set.

\begin{table}[ht]
%\small
    \centering
     \resizebox{.46\textwidth}{!}{
    \begin{tabular}{l|c|c|c}
        Model & P & R & F1 \\ \hline
        SemSynGTN & 70.2 & 57.2 & \textbf{63.0} \\ \hline
        %SemSynGTN - GTN & 66.0 & 52.5 & 60.3 \\
        SemSynGTN - GTN & 66.0 & 55.5 & 60.3 \\
        %SemSynGTN - Multi-hop & 68.1 & 56.7 & 61.0 \\ \hline
        SemSynGTN - Multi-hop & 68.1 & 55.2 & 61.0 \\ \hline
        %%SemSynGTN - Customization & 65.9 & 55.2 & 60.7 \\
        %SemSynGTN - SynCustom & 62.5 & 57.8 & 59.9 \\
        SemSynGTN - SynCustom & 62.5 & 57.5 & 59.9 \\
        SemSynGTN - SemCustom & 64.9 & 58.1 & 61.3 \\ \hline
        %SemSynGTN - SemCustom & 64.9 & 58.8 & 61.3 \\ \hline
        %%SemSynGTN - Diversity & 68.2 & 56.1 & 61.9 \\ 
        SemSynGTN - IB & 68.2 & 56.3 & 61.7 \\
        %SemSynGTN - IB + LSTM in $R$ & 67.5 & 55.3 & 61.1 \\
        SemSynGTN - IB + LSTM in $R$ & 67.5 & 55.8 & 61.1 \\
        %%SemSynGTN - Sem & 64.9 & 58.8 & 61.3 \\
        %%SemSynGTN - Syn & 62.5 & 57.8 & 59.9 \\
        %%SemSynGTN - Syn-Argument & 63.1 & 57.1 & 60.9 \\
        %%SemSynGTN - Syn-Trigger & 63.3 & 56.9 & 60.5
    \end{tabular}
     }
    \caption{The ablation study on the ACE 2005 dev set.}
    \label{tab:ablation}
\end{table}

The table clearly shows that all the components are necessary for SemSynGTN to achieve the highest performance. In particular,
the GTN model, the syntactic and semantic structure customization, and the structure multiplication are all important as eliminating any of them would hurt the performance significantly. These evidences highlight the importance of combining the customized sentence structures for EAE in this work. In addition, ``SemSynGTN - Bottleneck'' and ``SemSynGTN - IB + LSTM in $R$'' are also significantly worse than SemSynGTN, suggesting the effectiveness of information bottleneck to regularize the model for better generalization for GTNs in EAE.

{\bf Structure Analysis}: The proposed model generates four initial sentence structures in $\mathcal{A}$ (i.e., $A^d$, $A^a$, $A^e$, and $A^s$) to capture the general and task-specific structures for EAE based on the syntactic and semantic information. In order to evaluate their contribution for SemSynGTN, Table \ref{tab:view} presents the performance of the remaining models when each of these structures is eliminated from the model (i.e., from $\mathcal{A}$). It is clear from the table that the model performance is significantly worse when we remove any of the initial structures in $\mathcal{A}$, thus demonstrating the benefits of such structures for EAE. 

\begin{table}[ht]
%\small
    \centering
     \resizebox{.35\textwidth}{!}{
    \begin{tabular}{l|c|c|c}
        Model & P & R & F1 \\ \hline
        SemSynGTN & \textbf{70.2} & \textbf{57.2} & \textbf{63.0} \\ \hline
        %SemSynGTN - $A^d$ & 68.3 & 56.9 & 61.8 \\ 
        SemSynGTN - $A^d$ & 68.3 & 56.4 & 61.8 \\ 
        %%- Dep$^+$ & 68.0 & 58.7 & 62.0 \\
        SemSynGTN - $A^a$ & 65.2 & 57.1 & 60.9 \\
        %SemSynGTN - $A^a$ & 63.1 & 57.1 & 60.9 \\
        SemSynGTN - $A^e$ & 64.6 & 56.9 & 60.5 \\
        %SemSynGTN - $A^e$ & 63.3 & 56.9 & 60.5 \\
        SemSynGTN - $A^s$ & 65.2 & 55.9 & 60.2 \\
        %SemSynGTN - $A^s$ & 66.9 & 55.9 & 60.2 \\
    \end{tabular}
     }
    \caption{The structure contribution on the ACE 2005 dev set.}
    \label{tab:view}
\end{table}

{\bf Information Bottleneck Analysis}: In order to prevent overfitting for the GTN model in this work, we propose to cast GTN as an information bottleneck that seeks to minimize the mutual information between the GTN-produced vectors $H' = h'_1,h'_2,\ldots,h'_N$ and the representation vectors of the words from the earlier layers of the model (i.e., prior to GTN for the minimality of the representations). In particular, in the implementation of this idea, we propose to achieve the minimality of the representations by minimizing the mutual information between the vectors in $H'$ and the BiLSTM hidden vectors $H = h_1, h_2, \ldots, h_N$ in the sentence encoding. However, there are other prior layers of GTN whose hidden vectors can also be used for this MI minimization, including (1) the BERT-generated vectors for the words in the input sentence (i.e., $E = e_1,\ldots,e_N$ where $e_i$ is the hidden vector of the first wordpiece of $w_i$ in the last layer of the BERT model), and (2) the input vectors $X = x_1,\ldots,x_N$ for the BiLSTM layer in the sentence encoding where $x_i$ is the concatenation of $e_i$ and the relative distance embeddings for $w_i$ toward $w_a$ and $w_e$. In this analysis, we aim to evaluate the performance of SemSynGTN when the BiLSTM vectors $H$ are replaced by the vectors in $E$ and $X$ in the computation of $\mathcal{L}_{disc}$ for the MI minimization. Table \ref{tab:IB} presents the performance of these variants of SemSynGTN on the ACE 2005 development dataset.

\begin{table}[ht]
    \centering
    \resizebox{.48\textwidth}{!}{
    \begin{tabular}{l|c|c|c}
        Model & P & R & F1 \\ \hline
        SemSynGTN with $H$ for MI (proposed) & 70.2 & 57.2 & \textbf{63.0} \\ \hline
        SemSynGTN with $X$ for MI (BERT + distance) & 69.1 & 56.6 & 62.2 \\ 
        SemSynGTN with $E$ for MI (BERT) & 67.4 & 55.7 & 61.0 \\
    \end{tabular}
    }
    \caption{Performance of the models on the ACE 2005 development set using different configurations for MI in the information bottleneck.}
    \label{tab:IB}
\end{table}

The first observation from the table is that SemSynGTN with the vectors in $X$ for MI performs better than those with the BERT-generated vectors $E$. We attribute this to the fact that in addition to the BERT-generated vectors in $E$, the vectors in $X$ also include the relative distance embeddings of the words (i.e., for the argument candidate and trigger word). In principle, this makes $X$ more compatible with $H'$ than $E$ as both $X$ and $H'$ have access to the relative distances of the words to capture the positions of the argument candidate and trigger word in the sentences. Such compatible nature of information sources enables more meaningful comparison between $X$ and $H'$ for the MI minimization, providing more effective training signals to improve the representation vectors for EAE. More importantly, we see that the proposed MI minimization mechanism between $H'$ and $H$ helps SemSynGTN to achieve significantly better performance than those with the other variants (i.e., with $X$ or $E$ for the MI). This clearly helps to justify our proposal of employing the BiLSTM hidden vectors $H$ to compute $\mathcal{L}_{disc}$ in this work. In fact, the advantage of $H$ over $X$ for the MI minimization demonstrates the benefits of the BiLSTM layer to better combine the BERT-generated vectors $e_i$ and the relative distance embeddings for $w_i$ in $x_i$ to generate effective hidden vectors for the MI-based comparisons with $H'$ for EAE.

{\bf Performance Analysis}: To understand how the proposed model improves the performance over the baselines, we examine the outputs of SemSynGTN and the two major baseline models, i.e., (1) HMEAE \cite{wang2019hmeae}, the most related work that ignores the syntactic and semantic structures and previously has the best BERT-based performance for EAE, and (2) SemSynGTN - GTN that considers the syntactic and semantic structures but does not model their interactions to capture multi-hop paths with GTN. Our investigation suggests that while SemSynGTN outperforms HMEAE and SemSynGTN - GTN in general, the performance gaps between the models become substantially larger for the sentences with large numbers of words (i.e., distances) between the argument candidates and event triggers (called $\#bw$). In particular, Table \ref{tab:len} presents the performance of the three models on two subsets of the ACE 2005 development set, i.e., one with $\#bw \le 10$ and one with $\#bw > 10$. As we can see, the performance gaps between SemSynGTN and the two baselines on the subset with $\#bw > 10$ are much larger than those with $\#bw \le 10$. We attribute this better performance of SemSynGTN to its abilities to employ the combined structures based on syntax and semantic, and to model the multi-hop paths between words to compute the importance scores in the final structures of GTN. These abilities essentially enable SemSynGTN to capture longer and more flexible paths between words to compute effective representations for EAE. SemSynGCN is then able to perform better for the sentences with large $\#bw$ where encoding more context words is necessary to achieve high performance.

%for different sentences with different number of words between the target and the event trigger, we compare the performance of our model with two baselines on ACE 2005 Dev set. These baselines are (1) HMEAE \cite{wang2019hmeae} which ignores the syntactic and semantic structure and (2) \textbf{GTN/wo} which employs the syntactic and semantic structure but does not model their interactions. To this end, we group sentences based on the number of words between the argument and the trigger word into two intervals $\leq$10 words and $>$10 words. The results are shown in Table \ref{tab:len}. This table shows that while our model outperforms the two baselines for sentences with both $\leq$10 words and $>$10 words between the argument and the trigger word, the performance gap is larger for sentences with $>$10 words in between. This larger improvement for sentences with larger distance between the two words of interests shows the importance of utilizing the sentence structure for longer sentences. Moreover, from this table we observe that the performance gap between our model and HMEAE baseline is larger than our model and GTN/wo baseline. Our hypothesis is that the GTN/wo is still using the syntactic and the semantic structure while HMEAE completely ignores this. So GTN/wo has better performance in longer sentences than HMEAE baseline. 

\begin{table}[ht]
%\small
    \centering
    \resizebox{.40\textwidth}{!}{
    \begin{tabular}{l|c|c}
        Model & $\#bw \le 10$ & $\#bw > 10$ \\ \hline
        \textbf{SemSynGTN} & \textbf{64.1} & \textbf{61.3} \\ \hline
        SemSynGTN - GTN & 61.8 & 55.6 \\
        HMEAE \shortcite{wang2019hmeae} & 60.9 & 53.3 \\
    \end{tabular}
    }
    \caption{F1 scores of the models on the ACE 2005 dev set.} %based on the number of words between the argument candidate and the event trigger.
    \label{tab:len}
\end{table}

\section{Conclusion}

%We propose a novel deep learning model for the problem of event argument extraction that exploits the syntactic and semantic structures of the input sentences to learn effective representation vectors. The proposed model features the effective combination of the semantic and syntactic structures to encapsulate the multi-hop paths between the words for the importance score computation of the structures via Graph Transformer Networks. We also propose two novel inductive biases to improve the performance for EAE based on the structure diversity and information bottleneck techniques. The extensive experiments on two popular datasets for EAE demonstrate the effectiveness of the proposed model. In the future, we plan to apply the proposed model to other tasks in information extraction such as relation extraction and the similar tasks.

We propose a novel deep learning model for EAE that combines the syntactic and semantic structures of the input sentences for effective representation learning. The proposed model introduces Graph Transformer Networks and Graph Convolutional Networks to effectively perform the structure combination. A novel inductive bias is presented to improve the model generalization based on information bottleneck. The extensive experiments demonstrate the effectiveness of the proposed model. In the future, we plan to apply the proposed model to other related tasks in information extraction such as relation extraction.

%Event argument extraction (EAE) is one of the important sub-tasks of event extraction (EE) which has been relatively less studied than the other sub-tasks of EE. In this work, we proposed a new deep learning model capable of utilizing different structure of the sentence to find the role of an argument in a given event. Our model enjoys the new proposed architecture graph transfer network (GTN) and further addresses some of the limitations of the original architecture to efficiently combine different structural views of the sentence. Moreover, we proposed a new inductive bias in our model based on the information bottleneck principle to avoid overfitting. Our extensive experiments and analysis prove the effectiveness of the proposed model and establish new state-of-the-art results on two benchmark datasets for EAE. In future, we will explore this model for the other tasks of event extraction including event factuality.

\section*{Acknowledgement}

%This research has been supported in part by Vingroup Innovation Foundation (VINIF) in project code VINIF.2019.DA18 and Adobe Research Gift. This research is also based upon work supported in part by the Office of the Director of National Intelligence (ODNI), Intelligence Advanced Research Projects Activity (IARPA), via IARPA Contract No. 2019-19051600006 under the Better Extraction from Text Towards Enhanced Retrieval (BETTER) Program. The views and conclusions contained herein are those of the authors and should not be interpreted as necessarily representing the official policies, either expressed or implied, of ODNI, IARPA, the Department of Defense, or the U.S. Government. The U.S. Government is authorized to reproduce and distribute reprints for governmental purposes notwithstanding any copyright annotation therein. This document does not contain technology or technical data controlled under either the U.S. International Traffic in Arms Regulations or the U.S. Export Administration Regulations.

This research is based upon work supported in part by the Office of the Director of National Intelligence (ODNI), Intelligence Advanced Research Projects Activity (IARPA), via IARPA Contract No. 2019-19051600006 under the Better Extraction from Text Towards Enhanced Retrieval (BETTER) Program. The views and conclusions contained herein are those of the authors and should not be interpreted as necessarily representing the official policies, either expressed or implied, of ODNI, IARPA, the Department of Defense, or the U.S. Government. The U.S. Government is authorized to reproduce and distribute reprints for governmental purposes notwithstanding any copyright annotation therein. This document does not contain technology or technical data controlled under either the U.S. International Traffic in Arms Regulations or the U.S. Export Administration Regulations.

\bibliography{emnlp2020}
\bibliographystyle{acl_natbib}

\end{document}